%% file: neurips_2020.tex
\UseRawInputEncoding
\documentclass{article}

\PassOptionsToPackage{numbers, square}{natbib}
\usepackage[final]{orlr_neurips_2020}
\usepackage{enumitem,kantlipsum}

\usepackage[utf8]{inputenc} 
\usepackage[T1]{fontenc}    
\usepackage{hyperref}       
\usepackage{url}            
\usepackage{booktabs}       
\usepackage{amsfonts}       
\usepackage{nicefrac}       
\usepackage{microtype}      
\usepackage{amssymb}
\usepackage{amsmath}
\usepackage{graphicx}
\usepackage{wrapfig}

\title{Odd-One-Out Representation Learning}

\vspace{-1.2cm}
\author{%
  Salman Mohammadi, Anders Kirk Uhrenholt and Bj\o rn Sand Jensen\\
  School of Computing Science, \\University of Glasgow, Scotland\\
  \texttt{\small{salman.mohammadi@outlook.com},}
  \texttt{\small{a.uhrenholt.1@research.gla.ac.uk},}\\
  \texttt{\small{bjorn.jensen@glasgow.ac.uk}}\\
}
\graphicspath{ {./images/} }
\begin{document}

\maketitle

\vspace{-0.7cm}
\begin{abstract}
\vspace{-0.4cm}
    
  The effective application of representation learning to real-world problems requires both techniques for learning useful representations, and also robust ways to evaluate properties of representations. Recent work in disentangled representation learning has shown that unsupervised representation learning approaches rely on fully supervised disentanglement metrics, which assume access to labels for ground-truth factors of variation. In many real-world cases ground-truth factors are expensive to collect, or difficult to model, such as for perception. Here we empirically show that a weakly-supervised downstream task based on odd-one-out observations is suitable for model selection by observing high correlation 
  %
  on a difficult downstream abstract visual reasoning task. We also show that a bespoke metric-learning VAE model which performs highly on this task also out-performs other standard unsupervised and a weakly-supervised disentanglement model across several metrics.
\end{abstract}



\vspace{-0.25cm}
\section{Introduction}
\vspace{-0.2cm}

The ability to solve many different tasks in a variety of different environments is considered a hallmark of intelligence \cite{Cattell1963, Unsworth2014, Duncker1945}. Generalising in this manner requires representations of the world which effectively capture the most salient features of their environments \cite{Ha2018, Bengio2012a, Locatello2018b, VanSteenkiste2019a}. Recent work in representation learning has been focused on the learning and evaluation of \emph{disentangled} representations \cite{Higgins2016a, Ridgeway2018, Kumar2017, Kim2018, Eastwood2018a}. While such representations have been empirically shown to be useful for several downstream tasks \cite{Ha2018, Bengio2012a, Whitney2016, Achille2018}, unsupervised disentanglement techniques rely on disentanglement metrics that require full knowledge of underlying ground-truth factors of variation. 

While \citet{Locatello2019a} demonstrate that only observing very few fully labelled examples is sufficient for calculating disentanglement metrics, a number of interesting datasets will not have explicit labels for ground-truth factors and can therefore not be evaluated under existing metrics. For example, \citet{Locatello2020a} recently showed that comparison-based information (such as between ordered and temporally close video frames) can be used to learn disentangled representations. 
Without a particular focus on disentanglement, we extend previous work and demonstrate that similarity-based information is both relevant for learning representations but also for measuring useful properties of representations, where such measurements are otherwise inaccessible in many useful real-world scenarios. We contrast this approach with disentanglement learning and evaluation techniques to highlight the competitiveness of our method. 

In particular, we propose a metric which relies on implicit pairwise-comparison information from triplets of observations, where one out of three objects is labelled as the odd-one-out (OoO) providing weak similarity constraints (i.e. one object is the most dis-similar to all other objects, and two objects are the most similar). 
Secondly, we leverage the OoO information to show that a bespoke Variational Auto-encoder (VAE) model outperforms existing unsupervised and a weakly-supervised state-of-the-art disentanglement models across all disentanglement metrics (including the triplet-based metric as expected). 
Finally, we explore how useful weakly-supervised representations are for a practical downstream abstract visual reasoning task, using the procedurally generated Raven's Progressive Matrices (RPM) \cite{raven1938raven} task proposed by \citet{VanSteenkiste2019a}, which requires learners to infer abstract relations between objects and generalize these relations to unseen settings. 

\vspace{-0.1cm} 
\section{Methodology}
\vspace{-0.25cm}

We consider recent representation learning techniques which are largely based on Variational Auto-Encoders (VAEs) \cite{Kingma2013}, where the key underlying assumption is that datapoint $x$ is sampled from some stochastic process $p_\theta(x|z)$, dependent on a low-dimensional random variable $z$, considered to be the \emph{generative factors} of the data. State of the art unsupervised representation learning techniques aim to disentangle representations by regularizing the VAE objective. However, \citet{Locatello2018b} recently showed that these methods were dependent on random initial seed, hyperparameter selection, and, in particular, supervised disentanglement metrics which require full knowledge of the underlying factors of variation for model selection.

\begin{wrapfigure}{R}{0.12\textwidth}
\vspace{-0.6cm}
\includegraphics[width=0.12\textwidth]{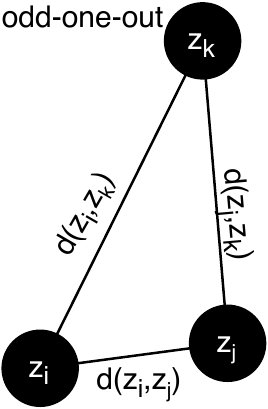}
\vspace{-0.99cm}
\label{fig:train_acc}
\end{wrapfigure}

\textbf{OoO Triplet VAE (TVAE):} We propose a VAE variant for an OoO paradigm which follows a common metric learning technique for learning semantically meaningful embeddings of perceptual information \cite{Agarwal2007, Tamuz2011, Ishfaq2018}. Such approaches rely on weak supervision settings such as similarity information between multiple observations. 

If the set of all individual objects is indexed by $i,j,k$, then a triplet observation, $m$, is a set $x^{(m)}_{ijk}=\{x_i,x_j,x_k\}$ where $x_k$ is known to be the odd-one-out thus implicitly providing the weak supervision. For each triplet observation, we embed the three objects in the latent space and assume it is a metric space such that we can interpret the odd-one-out supervision as providing two pairwise distance constraints as follows: since $k$ is the odd-one-out then the distance between $i$ and $k$, $d(z_i,z_k)$, should be larger than $d(z_i,z_j)$, and $d(z_j,z_k)$ larger than $d(z_i,z_j)$. We use the Euclidean distance evaluated based on the mean of $q_{\phi}(z|x)$. Each distance constraint is modeled as a Bernoulli random variable, $c$, indicating if the constraint is obeyed or not, and with the probability of the outcome defined by a Probit function, i.e., in general $p( c_{ {d(z_r,z_s) > d(z_r,z_t)} } = 1 \,|\, d(z_r,z_s), d(z_r,z_t)) = \Phi( d(z_r,z_s)^2 -  d(z_r,z_t)^2)$. Combining the two constrains allows us to model the triplet as a Bernoulli random variable , $y_k\in\{0,1\}$, encoding if $k$ is indeed the OoO or not based on the model with a proper likelihood defined as the product of two Probit terms, i.e., $p( y_k = 1 | z_i,z_j,z_k)= \Phi\left(d(z_i, z_k)^2 - d(z_i, z_j)^2\right) \cdot \Phi\left(d(z_j, z_k)^2 - d(z_i, z_j)^2\right)$ \footnote{We note that is it straightforward to extend this one-vs-all likelihood to a multinomial setting thus modelling which of three objects is the odd-one-out, however, this is not pertinent to the current investigation.}. 

With the aim to simultaneously embed the triplets and obey the constraint defined by the OoO information, we extend the standard VAE objective \cite{Kingma2013} with our new likelihood which yields the following lower bound on the marginal log-likelihood which should be maximised w.r.t $\theta$ and $\phi$:
\begin{equation}\begin{aligned}
\label{eqn:vae_loss}
    \mathcal{L}(\theta, \phi) = \mathbb{E}_{(x_i,x_j,x_k)}&\Big[\sum_{r\in{\{i,j,k\}}}\Big(\mathbb{E}_{q_\phi(z_r|x_r)}\left[\log p_\theta (x_r|z_r)\right] - D_{KL}(q_\phi(z_r|x_r)||p_\theta(z))\Big)\,+\\
    &\gamma\,\mathbb{E}_{q_\phi(z_i,z_j,z_k|x_i,x_j,x_k)}\left[ \log p(y_k|z_i,z_j,z_k )\right]\Big]\,,
\end{aligned}\end{equation}
%
%
where $\gamma$ varies the influence from the triplet loss on the (variational) posterior. While our approach is similar in spirit to \cite{Karaletsos2016, Ishfaq2018}, we emphasize that our formulation explicitly considers the odd-one-out paradigm resulting in two distinct pairwise distance constraints per triplet compared to their formulations only considering individual pairwise distance constraints. However, the additional information from a OoO observation can arguably be added as an additional albeit separate pairwise constraint in their methods. Moreover, in contrast to \citet{Karaletsos2016} we follow \citet{Locatello2020a} and jointly embed the three objects for each triplet along with the constraint instead of considering these separate aspects of the observation.

\textbf{Triplet Score:} To ensure fairness across a variety of possible representation learning techniques, and consistency with existing disentanglement metrics, we evaluate representations by using the predictive performance of a low-VC classifier identifying the odd-one-out from our generative model detailed in Appendix \ref{apx:triplet_model} (rather than just using the likelihood $p(y_k|z_i, z_j, z_k)$ in Eq.~\ref{eqn:vae_loss}). We define the \emph{Triplet Score} as the accuracy a low-VC classifier achieves in predicting the odd-one-out, given an encoding model $q_\phi(z|x)$, a number of points $m$, a classifier $clf$, as follows:
\vspace{-0.25cm}
\begin{enumerate}[leftmargin=0.3cm]
    \itemsep0em 
    \item[$\circ$]\textbf{Step 1}: Sample $m$ triplets from the triplet generative model.
    \item[$\circ$]\textbf{Step 2}: Compute $\mu_\phi(x)$ for each object in the triplet, for all $m$ triplets of observations, where $q_\phi(z|x) = \mathcal{N}(\mu_\phi(x), \sigma_\phi(x))$. Each $\mu_\phi(x)$ in the triplet is concatenated to create a single training point per triplet, $[\mu_\phi(x_i), \mu_\phi(x_j), \mu_\phi(x_k)]$, $\forall$ $m$.
    \item[$\circ$]\textbf{Step 3}: Calculate the accuracy of predicting the index of the odd-one-out, using $clf$.
\end{enumerate}

With the exception of the UDR metric \cite{duan2020unsupervised}, common techniques for evaluating representations for model selection rely on full knowledge of all the ground-truth factors of variation. We emphasize that the triplet score only requires knowledge of which observation is the odd-one-out. Furthermore, we aim to compare against the UDR metric in future work, however, we note that in comparison our proposed method requires far less computational resource.
In line with previous work \cite{Locatello2018b, VanSteenkiste2019a} we calculate this score by training a GBT classifier on 10,000 points and test on 5,000 points, however, we note that any classifier capable of implicitly comparing the distances is suitable (such as a MLP).

%
\section{Experiments}

\vspace{-0.2cm}

We build our experimental procedure on the work by \citet{VanSteenkiste2019a}. Our study evaluates the Triplet Score to determine how much it agrees with existing disentanglement metrics, as well as evaluating the TVAE across these disentanglement metrics \cite{Barrett2018}. We then evaluate the same set of disentanglement models on a downstream abstract reasoning task to determine the extent to which Triplet Score correlates with predictive performance on the task, as well as how the TVAE improves predictive performance on this task compared to state-of-the-art unsupervised models and a weakly-supervised model. Overall, we evaluate 60 VAEs and 90 WReNs. 

\vspace{-0.1cm}

\textbf{Experimental setup for evaluating representations:} To provide a fair comparison for the Triplet Score and the TVAE, we evaluate several state of the art representation learning models as baselines. Following notation from \citet{Tschannen2018}, we use two unsupervised methods which regularize the VAE objective in the form of 
$
     \mathcal{L}(\theta, \phi, x) = \mathbb{E}_{q_\phi(z|x)}[-\log p_\theta (x|z)] + \beta KL(q_\phi(z|x)||p_\theta(z)) + \tau KL(q_\phi(z|x)||\prod_j q_\phi(z_j|x))
$. 
The $\beta$-VAE \cite{Higgins2016a} restricts the information capacity of the encoder by increasing $\beta \geq 1$, with $\tau=0$. Similarly, the $\beta$-TCVAE \cite{Chen2019a} penalizes an estimate of the total correlation of the representation, with $\tau=1$ and $\beta > 1$. We also consider the group-based and weakly-supervised Ada-GVAE model of \citet{Locatello2020a}, which learns disentangled representations from pairs of observations that are similar to one another. We use the extension to the dSprites dataset \cite{dsprites17} proposed by \citet{VanSteenkiste2019a}, which adds object and background colour to the generative model. We  sample i.i.d. observations for the unsupervised models, and implement the paired-observations generative model in \citet{Locatello2020a} for the Ada-GVAE (we fix $k=rnd$ for this study). We implement our triplet generative model similar to the paired generative model which we detail in Appendix \ref{apx:triplet_model}. We plan to further evaluate our technique on the natural image THINGS \cite{Hebart2019} behavioural odd-one-out object dataset from \citet{Hebart2020}. 
 
We consider three different regularization strengths for each of the models, and train each model and hyperparamter choice five times with different random seeds, resulting in 60 representation learning models. We evaluate each model for the Factor VAE score \cite{Kim2018}, the B-VAE score \cite{Higgins2016a}, the DCI Disentanglement score \cite{Eastwood2018a}, and our Triplet Score. We detail our hyperparameters and implementation details in Appendix \ref{apx:setup}.
%
\begin{figure}[b!]
\centering
\includegraphics[width=\textwidth]{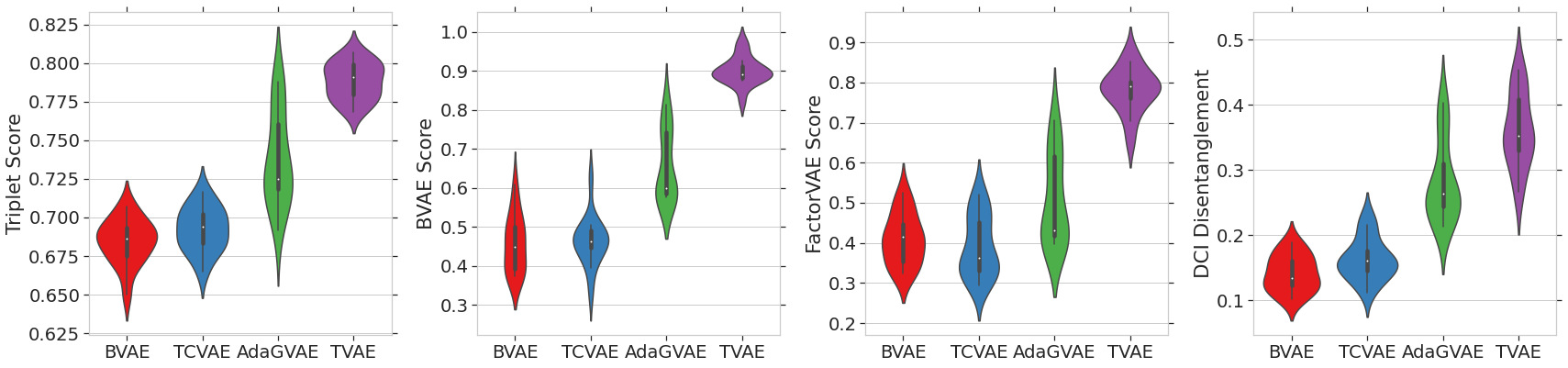}
\vspace{-0.4cm}
\caption{The TVAE outperforms all other models across the different disentanglement scores, and another weakly supervised model that uses comparison information between observations.}
\label{fig:disentanglement_scores}
\end{figure}

\textbf{Triplet-based weak supervision:} In Figure \ref{fig:disentanglement_scores}, we compare the different disentanglement metrics across the models. We observe that the weakly supervised models outperform the unsupervised models, and that the TVAE performs the best across all metrics. 


This result is expected considering the TVAE has access to weak labels which the other models do not. However we note that our modified TVAE objective function does not directly promote disentanglement, thus we leave an explanation of our empirical observations to future work. We also note that the Ada-GVAE, which is trained on a similar task based on the similarities between pairs of observations, outperforms the other unsupervised models on the Triplet Score. 



\begin{figure}[t]
\centering
\includegraphics[width=0.35\textwidth]{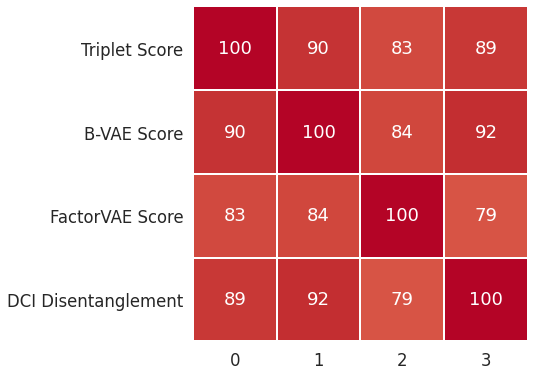}
\includegraphics[width=0.64\textwidth]{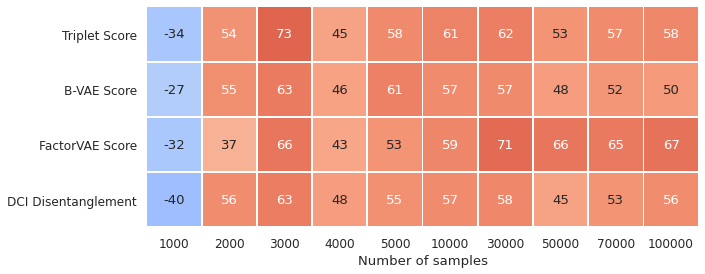}
 \vspace{-0.45cm}
\caption{(Left) Spearman's rank correlation between the different disentanglement metrics. (Right) Spearman's rank correlation between the disentanglement metrics and the downstream accuracy of the WReN during training, at different sample sizes.}
\label{fig:scores_cor}
\end{figure}

We provide the rank correlations between the different disentanglement scores in Figure \ref{fig:scores_cor} (Left), which show that the Triplet Score is very strongly correlated with existing disentanglement metrics. We note that this is likely due to the fact that VAE models tested happen to axis-align well, disentangling their observations by default, rather than indicating that our metric specifically captures disentanglement properties. 


\begin{wrapfigure}{R}{0.45\textwidth}
\vspace{-0.5cm}
\includegraphics[width=0.45\textwidth]{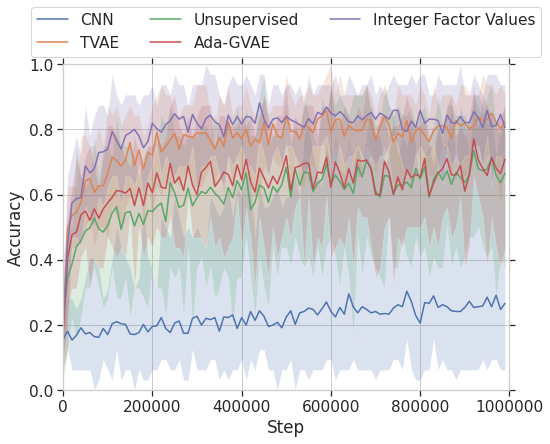}
\vspace{-0.7cm}
\caption{The TVAE improves sample efficiency and overall predictive performance on the RPM task.}
\vspace{-0.4cm}
\label{fig:train_acc}
\end{wrapfigure}

\textbf{Experimental setup for evaluating downstream abstract reasoning:}
In line with \citet{VanSteenkiste2019a}, we evaluate the sample efficiency of representations produced by disentanglement models above by using the Wild Relational Network (WReN) to solve a procedurally generated RPM task. Our study similarly investigates the predictive performance of WReNs on solving the RPM task at various sample sizes, and controlling for different sources of panel embeddings: integer-encoded ground-truth panel factor values, CNN embeddings trained from scratch, embeddings from the unsupervised disentanglement models, embeddings from the weakly-supervised Ada-GVAE, and embeddings from the TVAE. Further experimental details are outlined in Appendix \ref{apx:setup}.

\textbf{Abstract Reasoning:} In Figure \ref{fig:scores_cor} (Right) we show the rank correlation between the different disentanglement metrics and the WReN accuracy using embeddings from the representation learning models, at different sample sizes during training. We observe that the Triplet Score is equally highly correlated with all other disentanglement metrics across the sample sizes, indicating that the Triplet Score is also a reliable measure of how useful a representation is. Furthermore, In Figure \ref{fig:train_acc} we plot the training accuracy of the WReN during training using different sources of embeddings. We show that both weakly-supervised approaches outperform existing unsupervised approaches, and as expected, the TVAE outperforms all other approaches with the additional inductive bias from the weak supervision labels. 


\section{Conclusion}

\vspace{-0.2cm}

In this work, we proposed a simple weak supervision task for evaluating disentanglement in representation learning models based on odd-one-out triplet constraints. We also proposed a bespoke weakly-supervised representation learning model which is similar to existing metric-learning techniques, and empirically showed that the additional weak-supervision loss outperforms other unsupervised representation learning techniques. In our experiments, our weakly-supervised task was shown to correlate highly with predictive performance on an abstract visual reasoning task, without assuming any knowledge of ground-truth factors of variation. 


\textbf{Acknowledgements:}
\input{acknowledgement}

\bibliographystyle{unsrtnat}
\bibliography{neurips_2020_0}







\clearpage

\appendix
\section{Triplet Generative Model}\label{apx:triplet_model}

\begin{figure}[h]
\centering
\includegraphics[width=.32\textwidth]{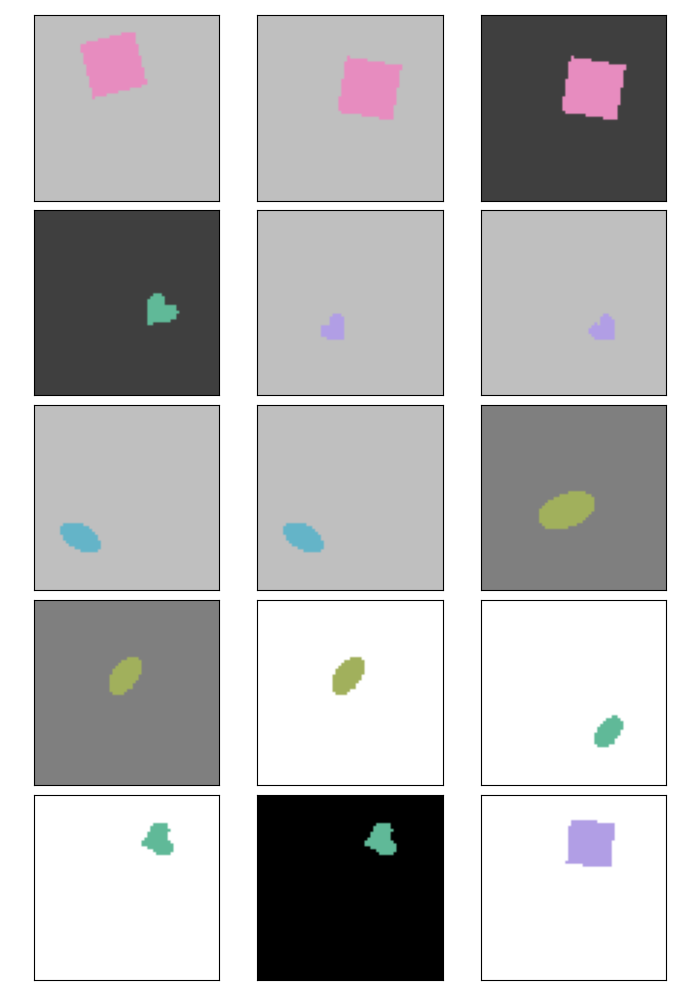}
\includegraphics[width=.32\textwidth]{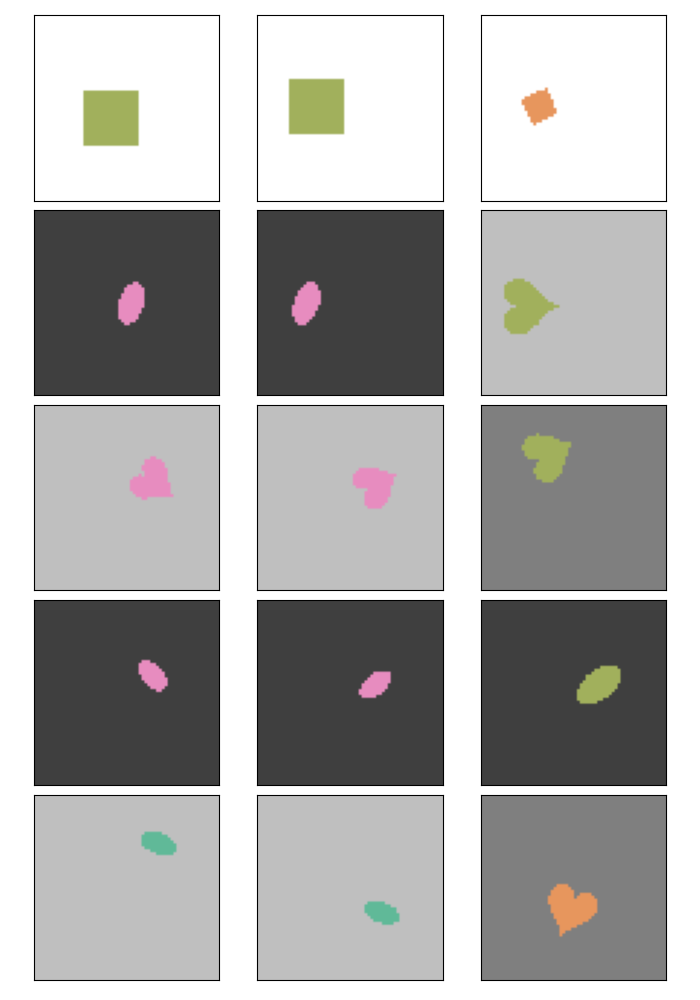}
\includegraphics[width=.32\textwidth]{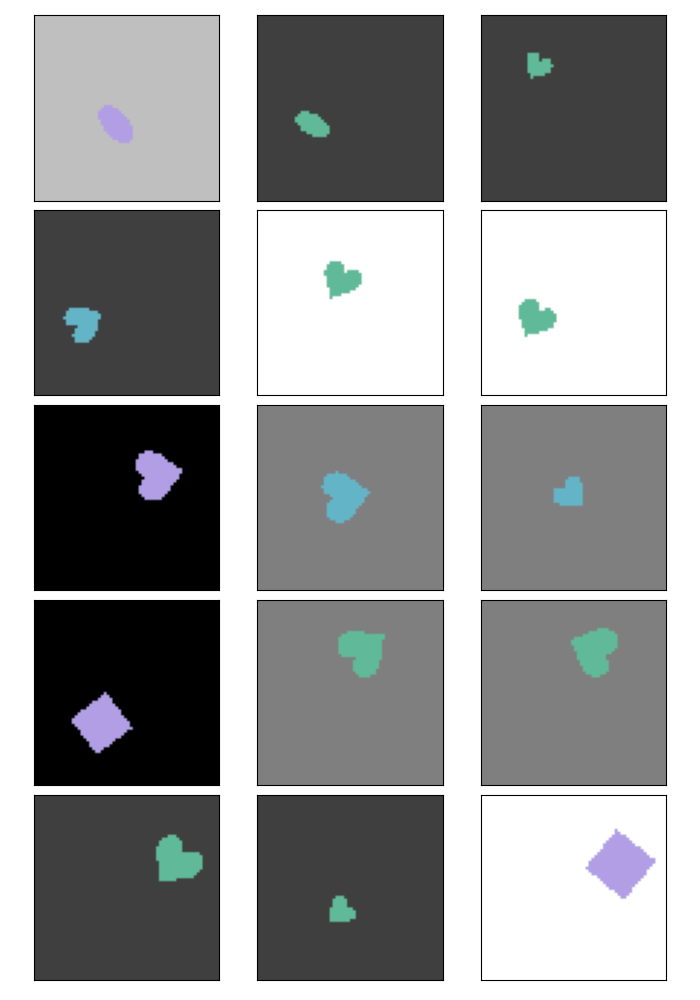}
\caption{Samples from the dSprites triplet dataset, with fixed $k=1$ (left), and $k\backsim unif(1,3)$ (center and right), for each row of triplets. We note that fixing $k=1$ makes the differences between the two most similar images, and the similarities between the odd-one-out, sparse, and easy to perceive, e.g. $x_1, x_2$ have only one factor difference, and $x_1, x_3$ have only one factor in common, whereas randomly sampling $k$, in the most difficult case $k=3$, results in $x_1, x_2$ having 3 factors difference, and $x_1, x_3$ having 3 factors in common.}
\label{fig:triplet_samples}
\end{figure}

Given a ground-truth model with $d$ i.i.d. generative factors $p(z)=\prod_i^d p(z_i)$, we aim to sample $z_1, z_2, z_3$ such that $z_1, z_2$ differ in exactly $k$ factors, $z_1, z_3$ differ in $d-k$ factors, and $z_2, z_3$ differ in $d$ factors. In this study, we consider randomly sampling $k\backsim unif(1, \lfloor\frac{d}{2}\rfloor)$ to respect the triplet constraint $d-k > k$, but invite further work to consider fixed $k$ to reflect real-world scenarios when the similarity conditions are known and consistent. We note that in the case of fixed $k=1$, the differences between the two most similar images, and the similarities between the odd-one-out, sparse, and easy to perceive, e.g. $x_1, x_2$ have only one factor difference, and $x_1, x_3$ have only one factor in common. We define our generative model as follows: 
\vspace{-0.1cm} 
\begin{equation}
\label{eqn:triplet_model}
    p(z_1)=\prod_{i=1}^D p(z_i), \qquad p(z_2)=\prod_{i=1}^k p(z_{S_i}), \qquad  p(z_3)=\prod_{i=1}^{D-k} p(z_{R_i}).
\end{equation}
Where we define $S=\{S\subset[d]:|S|=k\}$ as the subset of factor indices which will be re-sampled between $z_1$ and $z_2$, and $R=\{R\subset[d]:|R|=d-k\}$ as the subset of factor indices which will be re-sampled between $z_1$ and $z_3$, and other share factor values for factor indices which aren't in either set for each pair i.e. 
$
    p(z_{1_i}) = p(z_{2_i}),\, \forall i\notin S,\,\,\,\,\,  p(z_{1_i}) = p(z_{3_i}),\, \forall i\notin R.
$
We include samples from our triplet generative model in Figure \ref{fig:triplet_samples}.

\section{Architecture and Setup}\label{apx:setup}

We reuse all architectures, optimizers, batch size, and metric implementations from \citet{VanSteenkiste2019a}, and implement our study using the PyTorch framework \cite{NEURIPS2019_9015}. Similar to their approach, we also consider a sweep over a single hyperparameter for each model, and fix all others: $\beta \in [1, 6, 16]$ for $\beta$-VAE and Ada-GVAE, $\beta \in [2, 6, 16]$ for TCVAE. we consider $\gamma \in [1, 6, 16]$ for TVAE.

We use the abstract visual reasoning task proposed by \citet{VanSteenkiste2019a}, based on procedurally generating Raven's Progressive Matrices from ground-truth disentanglement models (we refer to their work for samples from the dataset). RPMs consist of three rows of context panels, with abstract relations present across each row, and the final row being incomplete. The task involves completing the final row by selecting from a set of possible answer panels, of which only one in correct. Selecting the correct answer panel requires the participant to infer the abstract relations across each row, and apply this knowledge to predict which answer panel correctly completes the final row.  Recent work by \citet{Barrett2018} explored the diagnostic ability of RPMs to measure abstract reasoning in neural networks. Using a dataset of procedurally generated RPMs and a variety of generalisation schemes, they found that models without significant inductive bias towards learning relational concepts perform poorly on the RPM task. 

For evaluating representations on the RPM task, we use the WReN \cite{Barrett2018}. The WReN was proposed as an extension to the original Relational Network (RN) module for solving RPMs by repeatedly applying the RN module on context and answer panels. Given a set of objects $O=\{o_1, o_2, ...,o_n\}, o_i\in\mathbb{R}^n$:
\begin{equation}
    RN(O) = f_\phi(\sum_{i, j} g_\theta(o_i, o_j))
\end{equation}
The RN infers relations between pairs of objects using neural networks $g_\theta$ and $f_\phi$, where $g_\theta$ determines how two objects are related, and $f_\phi$ scores this relation for the given task by considering all of the relations as a whole. 
Given $C = \{c_1, c_2, ..., c_8\}$ context panels, and $A = \{a_1, a_2, ..., a_6\}$ answer panels, the WReN produces a score $s_k$ for each answer panel $a_k$, by evaluating the RN over all pairs in the set of objects $E=\{C\}\cup\{a_k\}$. Intuitively, for a particular answer panel, we can think of this as the RN inferring the relations between every pair of context panels, and also between every context panel and the given answer panel, and then integrating information about all relations to produce a score for the answer panel. Formally, the WReN is evaluated as:
\begin{equation}
    \textrm{WReN}(C, A) = f_\phi\left(\sum_{x, y \in E}g_\theta(x, y)\right)
\end{equation}

In our study, when training embeddings for the WReN from scratch using a CNN, we use the implementation of \citet{VanSteenkiste2019a} over \citet{Barrett2018}, as the former authors don't use batch normalization after each convolutional layer. We evaluate the accuracy of the WReN on 100 newly sampled batches every 1000 steps.

\end{document}

%% file: acknowledgement.tex
SM acknowledge support from UKRI Centre for Doctoral Training in Socially Intelligent Artificial Agents supported by the Engineering and Physical Sciences Research Council (EPSRC) grant EP/S02266X/1. BSJ acknowledge support from EPSRC grant EP/R018634/1.